\documentclass{svmult}
\usepackage{mathptmx}
\usepackage{helvet}
\usepackage{courier}
\usepackage{amsmath}
\usepackage{amsfonts}
\usepackage{amssymb}
\usepackage{graphicx}
\usepackage{subfigure}
\usepackage[bottom]{footmisc}

\newcommand{\R}{\mathbb{R}}

\title*{Hidden cusps}

\author{Michel Coste, Philippe Wenger \and Damien Chablat}
\institute{%
  Michel Coste \at
  IRMAR, Universit\'e de Rennes 1 -- CNRS,
  \email{michel.coste@univ-rennes1.fr}\\
  Philippe Wenger \at
  IRCCyN, CNRS -- \'Ecole Centrale de Nantes,
  \email{philippe.wenger@irccyn.ec-nantes.fr}\\
  Damien Chablat \at 
  IRCCyN, CNRS -- \'Ecole Centrale de Nantes,
  \email{damien.chablat@irccyn.ec-nantes.fr}\\
  Authors partially supported by ANR-14-CE34-0008-01 Kapamat
  }

\begin{document}
\maketitle

\abstract{%
This paper investigates a situation pointed out in a recent paper, in which a non-singular change of assembly mode of a planar 2-RPR-PR parallel manipulator was realized by encircling a point of multiplicity 4. It is shown that this situation is, in fact, a non-generic one and gives rise to cusps under a small perturbation. Furthermore, we show that, for a large class of singularities of multiplicity 4, there are only two types of stable singularities occurring in a small perturbation: these two types are given by the complex square mapping and the quarto mapping. Incidentally, this paper confirms the fact that, generically, a local non-singular change of solution must be accomplished by encircling a cusp point.}

\keywords{parallel robot, cusp, non-generic singularity, perturbation}

\section{Introduction}

The non-singular change of assembly mode in parallel manipulators, first observed by C.~Innocenti and V.~Parenti-Castelli \cite{IPC}, is often associated with the presence of cusps and the non-singular change of assembly mode is realized by turning around a cusp point, or a cuspidal edge of the singularity surface (see for instance \cite{MAD,ZWC,CWC,HSCW}). It has also been reported that non-singular change of assembly modes can be realized by following an ``alpha curve'' (i.e. a fold curve intersecting itself transversally) \cite{BWS,MAPH}, and that the presence of cusps is not necessary for the existence of non-singular assembly mode changes \cite{CCW}.

A recent paper \cite{1} exhibits an example of a 2-dof parallel manipulator with an isolated singularity of multiplicity 4 of the inverse kinematics mapping, such that circling around the image of this singularity in the joint space results in a non-singular assembly mode change; moreover, after a second loop around the singularity, one is back in the same assembly mode. There is no cusp in the picture, but we intend to explain in the present paper that actually the cusps are hidden. Precisely, the singularity of multiplicity 4 is not a stable singularity, which means that it disappears under a small perturbation of the geometry of the manipulator, giving rise to three cusp points; in the joint space, the isolated singularity is perturbed into a deltoid curve with three cusps. Hence, circling around the singularity of multiplicity 4 was actually circling around 3 degenerate cusps.

H.~Whitney \cite{2} has shown that the only stable singularities of mappings between spaces of dimension 2 are folds and cusps. Any other higher order singularity becomes a combination of folds and cusps after perturbation, which amounts to say that these higher order singularities are degenerations of folds and cusps. We shall show that the case study of the perturbation of the 2-dof manipulator actually describes two main cases of singularities of multiplicity 4 (complex square and quarto mappings) leading to two different perturbations (the former with three cusps, the latter with one cusp).

\section{A case study}

\subsection{2R\underline{P}R-PR with higher order singularities}
\begin{figure}
\begin{center}
\includegraphics[scale=.5]{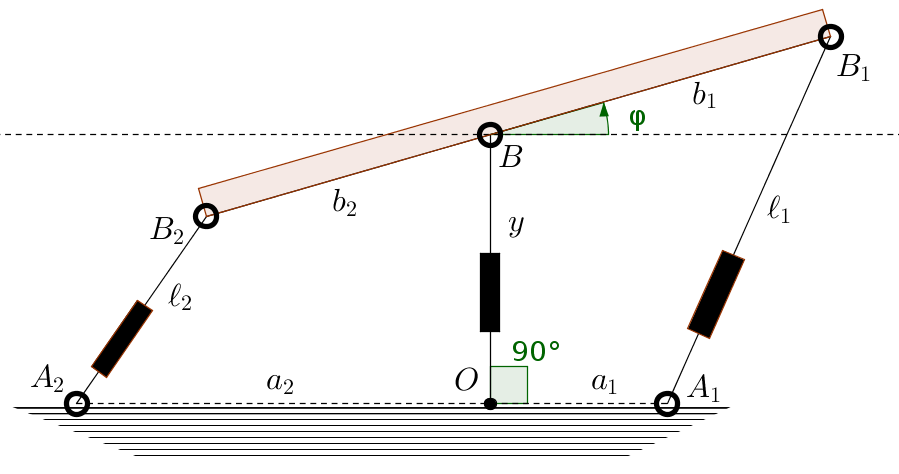}
\end{center}
\caption{Architecture of 2R\underline{P}R-PR\label{Fig1}}
\end{figure}

We begin with the example given in \cite{1}. It is a 2R\underline{P}R-PR planar manipulator with architecture described in Figure \ref{Fig1}.

The output coordinates are the angle $\varphi$ and the $y$-coordinate of the revolute joint $B$ which is constrained to move on the vertical axis. The input coordinates are the square $\ell_1^2$ and $\ell_2^2$ of the lengths of the legs $A_1B_1$ and $A_2B_2$. The equations for the inverse kinematic mapping are
\begin{equation}\label{IKM}\ell_i^2=y^2 - 2\,b_i\,y\,\sin(\varphi) + a_i^2-2\,a_i\,b_i\,\cos(\varphi) + b_i^2 \quad\text{for } i=1,2
\end{equation}
The Jacobian matrix $\mathrm{Jac}$ of the inverse kinematic mapping and its Jacobian determinant $J$ (up to a factor 4) are:
\begin{equation}\label{Jac}
\begin{aligned}
\mathrm{Jac}&=2\, \begin{pmatrix} -b_1\,y\,\cos(\varphi)+a_1\,b_1\,\sin(\varphi)&\quad y-b_1\,\sin(\varphi)\\
-b_2\,y\,\cos(\varphi)+a_2\,b_2\,\sin(\varphi)&\quad y-b_2\,\sin(\varphi)
\end{pmatrix}\\
J&= (b_1 + b_2)\,\cos(\varphi)\,y^2 + (a_1\,b_1 -
a_2\,b_2)\,\sin(\varphi)\,y-(a_1 + a_2)\,b_1\,b_2\,\sin(\varphi)^2 \;.
\end{aligned}
\end{equation} 
The possible cusp points and higher order singularities may be detected by adding to $J=0$ the equations 
\begin{equation}\label{cusp+}
\mathrm{Jac}\, \begin{pmatrix} -J_y\\ J_\varphi\end{pmatrix}=\begin{pmatrix} 0\\0\end{pmatrix}\;,
\end{equation}
where $J_\varphi$ and $J_y$ denote the partial derivatives of $J$ with respect to $\varphi$ and $y$. These equations express that the curve of singular points in the workspace either has a singularity or has a tangent vector in the kernel of the Jacobian matrix. We observe that $(\varphi,y)=(0\bmod \pi,0)$ satisfy $\mathrm{Jac}=$ the zero matrix (and hence also $J=0$ and equations (\ref{cusp+}) hold); these singularities are not cusps, but higher order singularities. 

\subsection{An example}

We compute the singularities in the workspace and in the joint space for an example with $a_1=3$, $a_2=7$, $b_1=6$, $b_2=5$, which is the same as the one considered in \cite{1}. In this case we can check that the only real solutions of $J=0$ and the equations (\ref{cusp+}) are $(\varphi,y)=(0\bmod \pi, 0)$.

For $(\varphi,y)=(0, 0)$, we have $\ell_1=3$, $\ell_2=2$. Developing the equations for the inverse kinematic mapping and for $J$ in a neighbourhood of $(0,0)$ we get
\begin{eqnarray}\label{dl00}
\ell_1^2-9&=&y^2+12\,y\,\varphi+18\,\varphi^2+\mathrm{h.o.t.}\;,\quad
\ell_2^2-4=y^2-10\,y\varphi+35\,\varphi^2+\mathrm{h.o.t.}\nonumber\\
J&=&11\,y^2-17\,y\,\varphi-300\,\varphi^2+\mathrm{h.o.t.}\;,
\end{eqnarray}
where h.o.t. stands for ``higher order terms''. This shows that the singularity is of multiplicity 4, and that the point $(\varphi=0,y=0)$ is a node of the curve of singularities (the discriminant $\Delta=(-17)^2-4\times11\times(-300)$ of the quadratic part of the development of $J$ at $(0,0)$ is positive).

For $(\varphi,y)=(\pi, 0)$, we have $\ell_1=9$, $\ell_2=12$. Developing $\ell_1^2-81$, $\ell_2^2-144$ and $J$ in a neighbourhood of $(\pi,0)$, with $\psi=\phi-\pi$, we get
\begin{eqnarray}\label{dlpi0}
\ell_1^2-81&=&y^2-12y\psi-18\psi^2+\mathrm{h.o.t.}\;,\quad
\ell_2^2-144=y^2+10y\psi-35\psi^2+\mathrm{h.o.t.}\nonumber\\
J&=&-11\,y^2+17\,y\,\psi-300\,\psi^2+\mathrm{h.o.t.}\;.
\end{eqnarray}
This shows that the singularity is also of multiplicity 4, and that the point $(\varphi=\pi,y=0)$ is an isolated double point of the curve of singularities (the discriminant $\Delta=17^2-4\times(-11)\times(-300)$ of the quadratic part of the development of $J$ at $(\pi,0)$ is negative).

\medskip

\begin{figure}[h!]
    \centering
    \subfigure[Workspace]
        {\begin{picture}(130,130)
        \put(0,0){\includegraphics[width=0.4\textwidth]{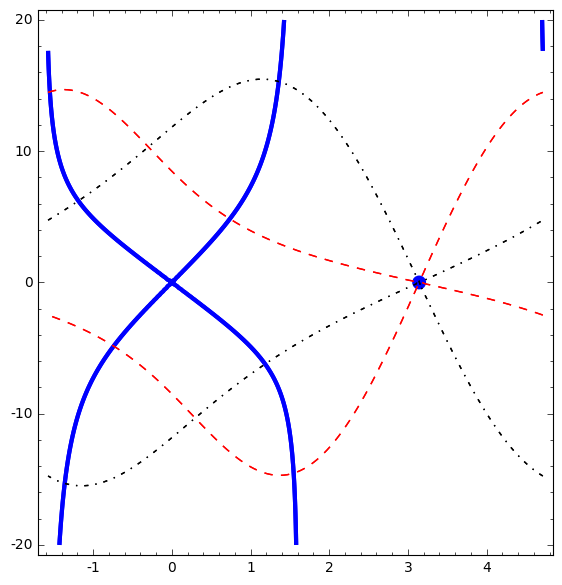} }
        \put(115,-5){$\varphi$} \put(-5,115){$y$}
        \put(65,67){1} \put(75,82){2} \put(93,92){3} \put(111,82){4}
        \put(121,67){1} \put(111,52){2} \put(93,42){3} \put(75,52){4}
        \put(33,82){1} \put(47,102){2} \put(33,115){3} \put(20,102){4}
        \put(41,52){1} \put(29,34){2} \put(41,19){3} \put(56,34){4}
        \put(19,67){1} \put(12,50){2}
        \end{picture}}  
    \quad\quad\quad
    \subfigure[Joint space]
        {\begin{picture}(120,120)
        \put(0,0){\includegraphics[width=0.4\textwidth]{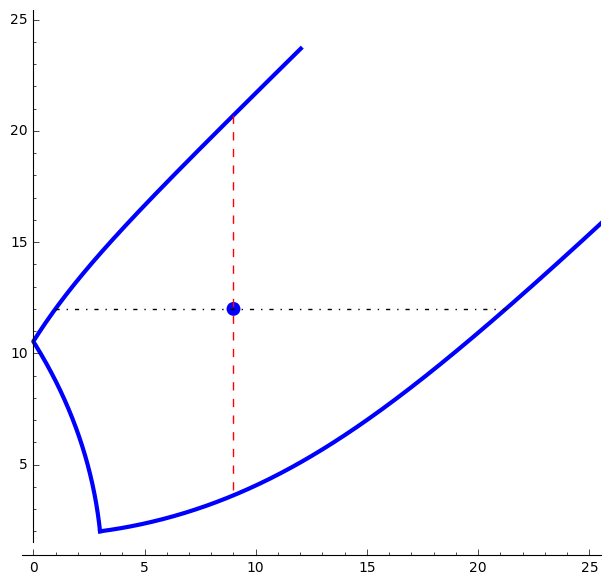} }
        \put(115,-5){$\ell_1$} \put(-5,115){$\ell_2$}
        \put(35,45){1} \put(65,45){2} \put(65,65){3} \put(35,65){4}
        \end{picture}}
    \caption{\label{wsjs}Workspace and joint space of the manipulator}
\end{figure}
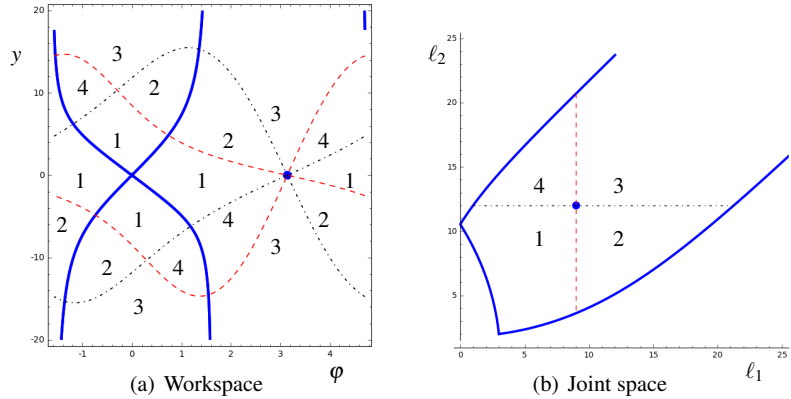

Figure \ref{wsjs} (a) represents the workspace of the manipulator; it must be understood that the right side ($\varphi=3\pi/2$) has to be identified with the left side ($\varphi=-\pi/2$). The singularity curve is represented in thick blue; one can see the node at $\varphi=0, y=0$ and the isolated double point at $\varphi=\pi, y=0$. The dash-dot black curve is the level curve $\ell_1=a_1+b_1=9$ and the dashed red curve is the level curve $\ell_2=a_2+b_2=12$. The numbers in the zones delimited by these curves indicate the corresponding images by the inverse kinematic mapping in the joint space.

The joint space is represented in Figure \ref{wsjs} (b) (the same figure appears in \cite{1}). One can see the image of the singularity curve in blue. Inside the domain delimited by this curve, the direct kinematic problem has four solutions, except at the image $(\ell_1=9,\ell_2=12)$ of the isolated singularity point where there is one solution of multiplicity 4 and two other solutions. Above each point of the image singularity curve, there are two double solutions, except at the point $\ell_1=3$, $\ell_2=2$ (image of the node) where there is one solution of multiplicity 4. The zones numbered 1,2,3,4 are the images of the zones with the corresponding numbers in the workspace. 

It can be seen that circling around the isolated singularity point in the joint space, following the numbering 1-2-3-4-1, yields a non-singular assembly mode change leading from a zone numbered 1 in the workspace touching the isolated singularity point to the other one. A second loop makes one return to the initial assembly mode. This phenomenon cannot be faithfully represented in a 3-dimensional reduced configuration space: one cannot have a ramp turning around the singular configuration and returning to the start level after two turns without an artificial self-intersection. 

One can see in this example a non-singular assembly mode change by circling around a singularity which is not a cusp. The hidden cusps are revealed by slightly perturbing the geometry of the manipulator.

\subsection{Revealing the hidden cusps}

\begin{figure}[h!]
\begin{center}
\includegraphics[scale=.5]{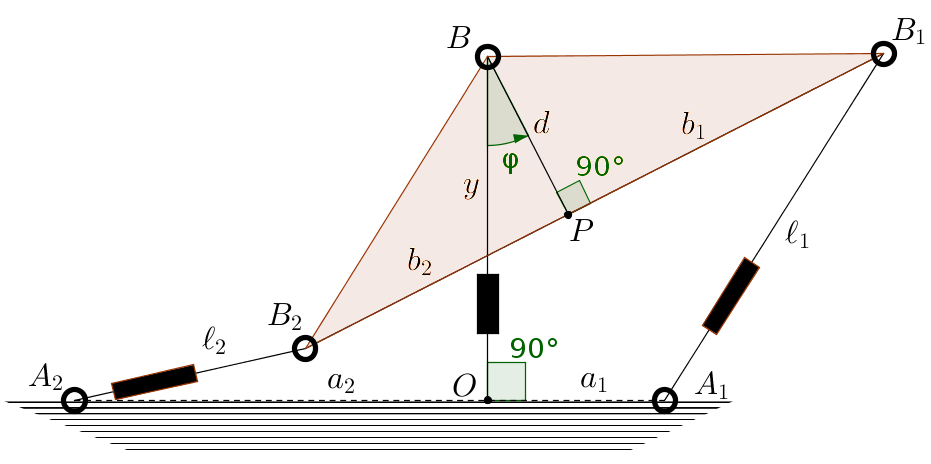}
\end{center}
\caption{\label{robpert}The modified manipulator}
\end{figure}

The manipulator is modified so that the revolute joint $B$ on the platform is no longer on the line $B_1B_2$, but at a distance $d$ from this line: see Figure \ref{robpert}.

We compute the example with $a_1=3$, $a_2=7$, $b_1=6$, $b_2=5$ and $d=3$. The equations for the inverse kinematic mapping are now
\begin{equation}\label{IKMpert}
\begin{aligned}
\ell_1^2 &=y^2 - 6y(\cos(\varphi) - 2\sin(\varphi)) - 36\cos(\varphi) - 18\sin(\varphi) + 54\\
\ell_2^2 &=y^2 - 2y(3\cos(\varphi) + 5\sin(\varphi)) - 70\cos(\varphi) + 42\sin(\varphi) + 83\;,
\end{aligned}
\end{equation}
and the Jacobian determinant is, up to a constant factor,
\begin{equation}
\begin{aligned}
J&=11\,y^2\,\cos(\varphi) - y\,(30\,\cos(\varphi) + 17\,\sin(\varphi) + 33)\\
&{\quad } + 390\,\cos(\varphi)^2 -
30\,\cos(\varphi)\,\sin(\varphi) - 300
\end{aligned}
\end{equation}
We can detect the cusps or higher order singularities by solving the system formed by $J=0$ and equations (\ref{cusp+}).  We get four real solutions, which are 
\begin{equation}
\begin{aligned}
(\varphi\simeq-0.0023, y\simeq2.9069),&\; (\varphi\simeq2.6492, y\simeq-2.2190),\\
(\varphi\simeq-2.7368, y\simeq-1.2968),&\; (\varphi\simeq3.0855,y\simeq2.6935)
\end{aligned}
\end{equation} 
It can be checked that all four points are actually cusp points.

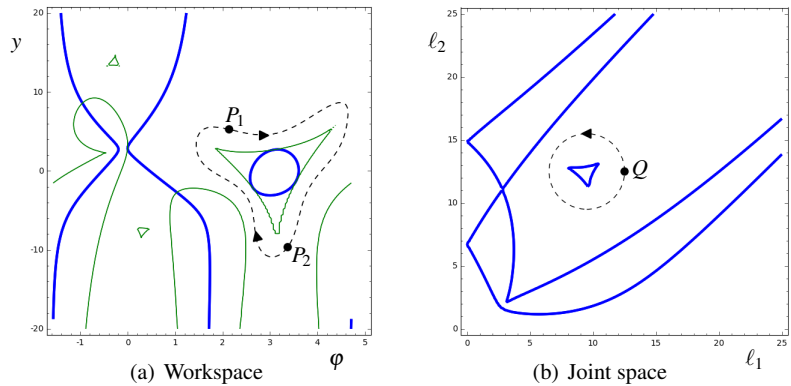
\begin{figure}[h!]
    \centering
    \subfigure[Workspace]
        {\begin{picture}(130,130)
        \put(0,0){\includegraphics[width=0.4\textwidth]{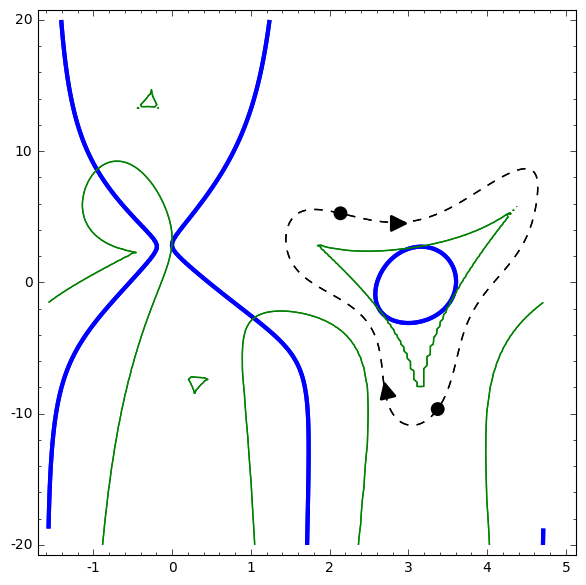} }
        \put(115,-5){$\varphi$} \put(-5,115){$y$}
        \put(76,87){$P_1$} \put(101,34){$P_2$}
        \end{picture}}  
    \quad\quad\quad
    \subfigure[Joint space]
        {\begin{picture}(120,120)
        \put(0,0){\includegraphics[width=0.4\textwidth]{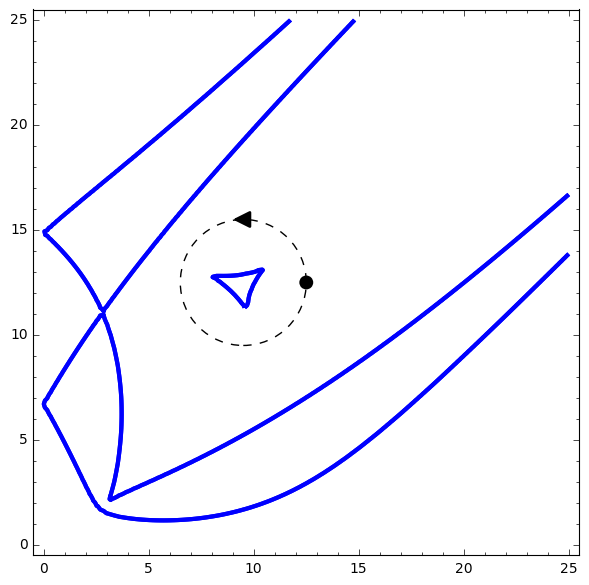} }
        \put(115,-5){$\ell_1$} \put(-5,115){$\ell_2$}
        \put(72,67){$Q$}
        \end{picture}}
\caption{\label{wsjspert}Workspace and joint space of the modified manipulator}
\end{figure}

The curve of singularities in the workspace is represented in thick blue in Figure \ref{wsjspert} (a). Note that it retains the overall features of the original singularity curve in Figure \ref{wsjs} (a), except for the node at $(0,0)$ which is simplified in two non-intersecting branches and the isolated double point at $(\pi,0)$ which has evolved into an oval. The characteristic curves in the workspace (defined in \cite{2bis}) are represented in green. One can recognize the cusp points as the points of tangency of the characteristic curves with the curve of singularities: there is one cusp point on one branch of the simplification of the node, and three cusp points on the oval obtained by perturbing the isolated double point of the singularity curve.
 
Figure \ref{wsjspert} (b) represents the joint space. One can see the four cusps, three on the central deltoid (image of the oval) and one on a branch of the outer curve (the two branches have also a crossing point). There are six solutions to the direct kinematic problem inside the deltoid, and 4, 2 or 0 solutions as one proceeds towards the outer regions.

Note that the connected zone in the workspace encircling the large green deltoid is not a uniqueness domain \cite{2bis}: it is a 2-sheeted covering of the zone around the deltoid in the joint space. This latter zone is not simply connected, so we cannot deduce that a connected component of its preimage is a uniqueness domain. 

The picture of the joint space shows that circling around the isolated singularity in the joint space was actually circling around three degenerate cusps. The dashed circle from $Q$ to $Q$ around the deltoid in the joint space lifts to the dashed trajectory from $P_1$ to $P_2$ in the workspace; a second turn on the circle completes the circuit from $P_2$ back to $P_1$.

\section{The general mathematical picture: unfolding of a singularity of multiplicity 4 of a mapping of surfaces}

We explain here how the observations made for the modified 2R\underline{P}R-PR fit into a general mathematical framework. We begin with by recalling two examples described in \cite {3}.

\subsection{Complex square mapping and its unfolding}

The first example is given by 
$f:(x,y)\longmapsto (u=x^2-y^2,\;v=2xy)$
which is the complex square function $z\mapsto z^2$, written in real and imaginary parts; this shows that every point in $\R^2$ is the image by $f$ of two points in $\R^2$, except the origin which is the image of the origin only. The Jacobian determinant of $f$ is, up to a constant factor, $x^2+y^2$. The only singularity of $f$ is at the origin, and this singularity has multiplicity 4 (the dimension of the quotient algebra $\R[x,y]/(x^2-y^2,2xy)$).

Now we perturb the mapping $f$ to 
\begin{equation}
\tilde f:(x,y)\longmapsto (u=x^2-y^2+4a x,\;v=2xy+4by)\;.
\end{equation}
The Jacobian determinant of $\tilde f$  becomes, up to a constant factor, $(x+a+b)^2+y^2-(a-b)^2$. If $b\neq a$, the set of singular points of $\tilde f$ is the circle with centre $(-a-b,0)$ and radius $|a-b|$. There are three cusp points on this circle, and the image curve in the $(u,v)$ plane is a deltoid with three cusps. A point inside the deltoid has four preimages, outside two. Circling around the deltoid permutes the two preimages (as circling around the origin does for the complex square root).

\begin{figure}[h!]
    \centering
    \subfigure[Perturbation of complex square]
        {\begin{picture}(140,70)\put(0,5){\includegraphics[width=0.2\textwidth]{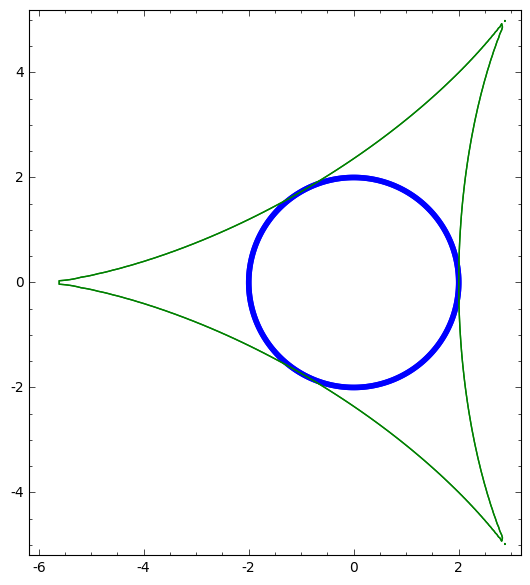}}
        \put(-5,72){$y$} \put(75,72){$v$} \put(65,0){$x$}
        \put(140,0){$u$}
        \put(80,5)
        {\includegraphics[width=0.2\textwidth]{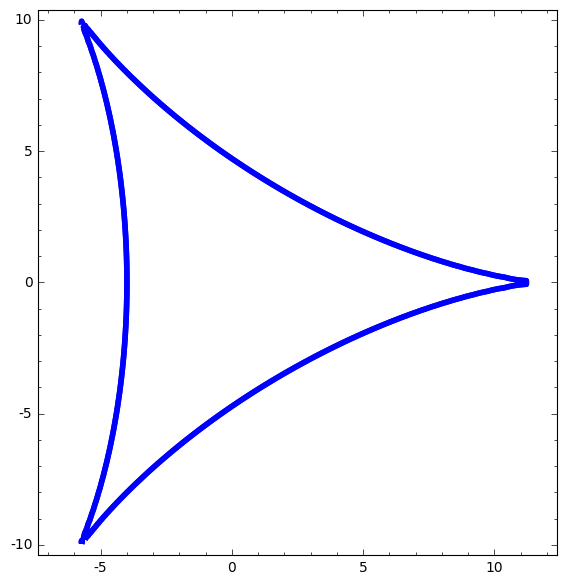} } \end{picture}}
    \qquad\qquad
    \subfigure[Perturbation of quarto]
        {\begin{picture}(140,70)\put(0,5){\includegraphics[width=0.2\textwidth]{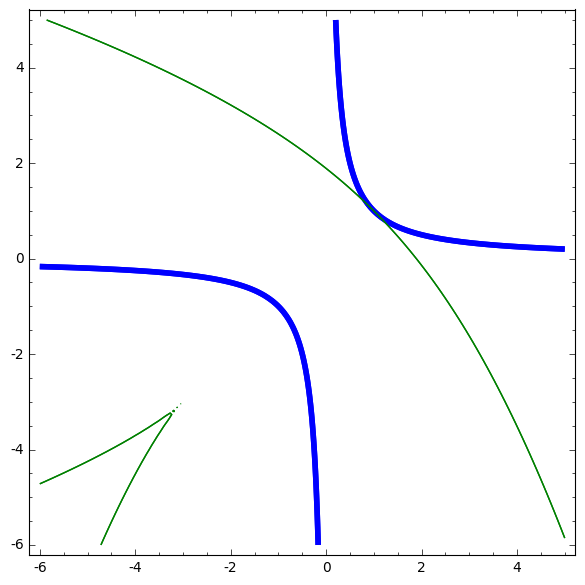}}
        \put(-5,72){$y$} \put(75,72){$v$} \put(65,0){$x$}
        \put(140,0){$u$}
        \put(80,5)
        {\includegraphics[width=0.2\textwidth]{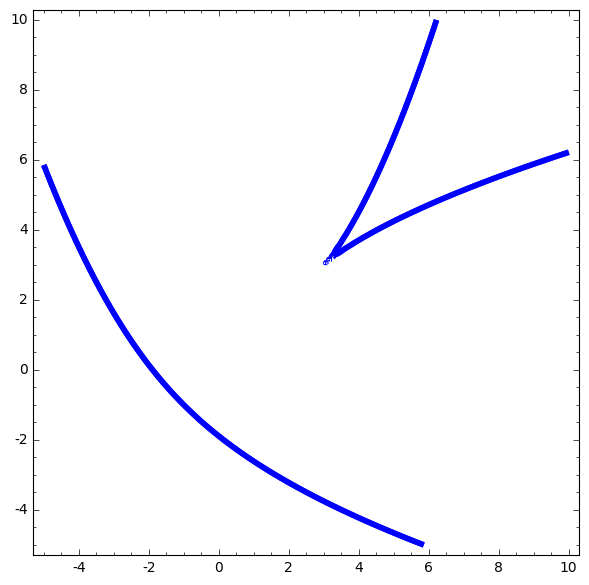} } \end{picture}}
    \caption{Perturbations of the complex square mapping and of the quarto mapping\label{squarequarto}}
\end{figure}
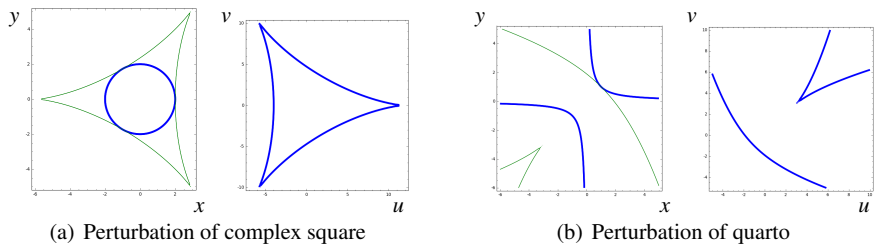

Figure \ref{squarequarto} (a) shows the situation at the source (coordinates $(x,y)$) and at the target (coordinates $(u,v)$), in the case $a=1$, $b=-1$. The blue curves are the curves of singularities. The green curve at the source is the characteristic curve; the cusps points are the points where the blue and green curves are tangent.
 
\subsection{Quarto mapping and its unfolding}
The second example is the mapping $g:(x,y)\longmapsto (u=x^2,\;v=y^2)$; this mapping is named ``quarto'' because it folds the plane $x,y)$ onto the first quadrant of the plane $(u,v)$, which is covered by four sheets.
The Jacobian determinant of $ g$ is, up to a constant factor, $xy$. The set of singular points of $g$ is the union of the two axes.

Let us perturb the mapping $g$ to
\begin{equation}
\tilde g:(x,y)\longmapsto (u=x^2+2ay,\;v=y^2+2bx)\;.
\end{equation}
The Jacobian determinant becomes, up to a constant factor, $xy-ab$. This is the equation of an equilateral hyperbola, if $ab\neq 0$. Its image by $\tilde g$ is a curve in the $(u,v)$ plane with two branches, one of which has a cusp; inside the cusp, each point has four preimages by $\tilde g$, between the branches two, and zero elsewhere.

Figure \ref{squarequarto} (b) represent the situation at the source and at the target in the case $a=b=1$  in the same way as for the preceding example. One can see the cusp point at the source.

\subsection{General case}

The two examples above are actually the complete list of the stable singularities that can be obtained by perturbing a singularity of multiplicity 4 where the $2\times 2$ Jacobian matrix is the zero matrix. These are the elliptic (complex square case) and hyperbolic (quarto case) $\Sigma^2$ singularities which are studied in \cite{AGV}, Part I \S3. The notation $\Sigma^2$ means that the Jacobian matrix has corank 2, i.e. is the zero matrix in dimension 2, and in this case multiplicity 4 is equivalent to the fact that the discriminant $\Delta$ of the quadratic part of the Taylor expansion of the Jacobian determinant at the singularity is nonzero. The elliptic case corresponds to $\Delta<0$ and the hyperbolic case to $\Delta>0$.

We can now return to the example of the 2R\underline{P}R-PR. We have $\Delta >0$ at the singularity $(\varphi,y)=(0,0)$ (see (\ref{dl00})): we are here in the case "quarto mapping" and we can clearly see the relevant parts of Figure \ref{wsjspert} corresponding to Figure \ref{squarequarto}(b). We have $\Delta<0$ at the singularity $(\varphi,y)=(\pi,0)$ (see (\ref{dlpi0})): we are now in the case "complex square mapping" and we can compare the relevant parts of Figure \ref{wsjspert} with Figure \ref{squarequarto}(a).

\section{Conclusion}

We have shown that the singularities of multiplicity 4 that appear in the study of the kinematics of the 2RPR-PR are not generic and give rise to cusps under a small perturbation. We have also shown that these singularities belong to a family of singularities which splits in two cases according to the sign of the discriminant of the quadratic part of the Jacobian determinant : the "complex square mapping" case and the "quarto mapping" case which are well known in the theory of singularities of differentiable mappings. In the first case, the singularity is isolated and circling around it in the joint space results in an exchange of two solutions to the direct kinematic problem. A small perturbation to stable singularities gives three cusp points which were in some sense "hidden" in the singularity of multiplicity 4, and so one can argue that this example does not invalidate the rule that, generically, local non-singular assembly mode changes  arise by circling around cusps. 

We have limited our study to the 2-dof case. In a future work, we shall examine the perturbation of the second example in \cite{1}, which is interesting because it gives a fully parallel generic 3-R\underline{P}R manipulator with properties similar to the ones we have seen for the constrained 2R\underline{P}R-PR. We shall also discuss in more details characteristic surfaces and uniqueness domains \cite{2bis}.

\end{document}